\documentclass[sigconf,natbib=true,anonymous=false]{acmart}
\usepackage{tabularx}

\AtBeginDocument{%
  }

\setcopyright{acmcopyright}
\copyrightyear{2025}
\acmYear{2025}
\setcopyright{acmlicensed}\acmConference[SIGIR '25]{Proceedings of the 48th
International ACM SIGIR Conference on Research and Development in
Information Retrieval}{July 13--18, 2025}{Padua, Italy}

\author{Yumin Kim}
\affiliation{
  \institution{Chung-Ang University}
  \city{Seoul}
  \country{South Korea}}
\email{kimym7801@cau.ac.kr}

\author{Hwanhee Lee}
\affiliation{
  \institution{Chung-Ang University}
  \city{Seoul}
  \country{South Korea}}
\email{hwanheelee@cau.ac.kr}

\acmISBN{978-1-4503-XXXX-X/18/06}
\acmDOI{XXXXXXX.XXXXXXX}

\begin{document}

\title{Selective Demonstration Retrieval for Improved Implicit Hate Speech Detection}

\begin{abstract}

Hate speech detection is a crucial area of research in natural language processing, essential for ensuring online community safety. However, detecting implicit hate speech, where harmful intent is conveyed in subtle or indirect ways, remains a major challenge. Unlike explicit hate speech, implicit expressions often depend on context, cultural subtleties, and hidden biases, making them more challenging to identify consistently. Additionally, the interpretation of such speech is influenced by external knowledge and demographic biases, resulting in varied detection results across different language models.  
Furthermore, Large Language Models (LLMs) often show heightened sensitivity to toxic language and references to vulnerable groups, which can lead to misclassifications. This over-sensitivity results in false positives—incorrectly identifying harmless statements as hateful—and false negatives—failing to detect genuinely harmful content. Addressing these issues requires methods that not only improve detection precision but also reduce model biases and enhance robustness. To address these challenges, we propose a novel method, which utilizes in-context learning without requiring model fine-tuning. By adaptively retrieving demonstrations that focus on similar groups or those with the highest similarity scores, our approach enhances contextual comprehension. Experimental results show that our method outperforms current state-of-the-art techniques. Implementation details and code are available at \url{TBD}.

\end{abstract}

\ccsdesc[500]{Information systems~Top-k retrieval in databases}
\ccsdesc[500]{Information systems~Clustering and classification}
\ccsdesc[500]{Information systems~Novelty in information retrieval}
\ccsdesc[500]{Computing methodologies~Natural language generation}
% \ccsdesc[500]{Computing methodologies~Information extraction}

\keywords{Retrieval-based In-Context Learning, Classification, Hate Speech Detection}



\maketitle

\section{INTRODUCTION}

The rapid growth of online media has led to increased exposure to hate speech, making the development of effective detection systems essential \cite{schmidt-wiegand-2017-survey}. 
While early research primarily focused on detecting explicit hate speech \cite{Mathew_Saha_Yimam_Biemann_Goyal_Mukherjee_2021}, recent studies have shifted attention toward implicit hate speech, which is often more widespread yet difficult to detect due to its subtle and indirect nature \cite{jurgens-etal-2019-just}.

%Implicit hate speech presents a significant challenge for natural language processing (NLP) systems, as it often relies on nuanced linguistic cues, cultural context, and indirect expressions that do not necessarily contain explicitly toxic words \cite{nielsen2002subtle}. 
%Due to this complexity, even the most advanced Large Language Models (LLMs) are unable to accurately distinguish between harmful and benign statements.
%Moreover, existing detection models frequently struggle with biases and over-sensitivity, leading to false positives when benign discussions involve sensitive topics or false negatives when implicit hate is subtly masked.
Implicit hate speech relies on nuanced linguistic cues, cultural context, and indirect expressions that often do not contain overtly toxic words \cite{nielsen2002subtle}. Due to this complexity, even the most advanced Large Language Models (LLMs) struggle to accurately distinguish between harmful and benign statements. Moreover, existing detection models frequently suffer from biases and over-sensitivity, leading to false positives when benign discussions involve sensitive topics and false negatives when hateful content is subtly masked.

Early approaches on hate speech detection predominantly relied on fine-tuning pre-trained models to improve classification accuracy~\cite{grondahl2018all, swamy-etal-2019-studying,  kim2022generalizable, yadav2024hatefusion}. With the advent of LLMs, in-context learning approaches, such as chain-of-thought (CoT) reasoning \cite{wei2022chain}, have been increasingly utilized in hate speech detection. For instance, \cite{yang-etal-2023-hare} explores fine-tuning rationales generated by LLMs through CoT reasoning, demonstrating improved interpretability in classification decisions. Additionally, few-shot learning performance has been enhanced by incorporating task decomposition strategies that predict the identities of targeted groups \cite{alkhamissi-etal-2022-token}. 

Despite these advancements, existing methods largely overlook two critical challenges: \textit{over-sensitivity} and \textit{dataset subjectivity}. Over-sensitivity in LLMs often results in models misclassifying neutral or even counter-hate statements as harmful, significantly reducing reliability \cite{zhang-etal-2024-dont-go}. Moreover, the inherent subjectivity of hate speech datasets can lead to inconsistencies in classification outcomes, further worsening bias-related issues \cite{masud-etal-2024-hate}.

To address these challenges, we introduce Adaptive Retrieval-based In-context Learning (ARIIHA), a novel framework designed to enhance implicit hate speech detection through adaptively retrieving demonstrations for in-context learning, without requiring additional fine-tuning \cite{huang-etal-2024-towards, hee-etal-2024-bridging}. 
%Our approach efficiently retrieves relevant demonstrations that align with both the linguistic patterns and the targeted group of the input, thereby improving contextual understanding and classification accuracy. By retrieving relevant demonstrations having similarly targeted groups, ARIIHA mitigates over-sensitivity by preventing the model from relying excessively on toxic words and reduces dataset subjectivity by dynamically adjusting to variations in annotation biases and dataset-specific decision patterns.
% ARIIHA leverages in-context learning by adaptively retrieving relevant demonstrations through two distinct methods: one prioritizes demonstrations that target similar groups, while the other selects demonstrations based on high similarity scores. Our approach dynamically selects between these retrieval strategies using thresholds based on similarity scores and the model's reliance on shortcut cues, ensuring that only robust contextual demonstrations inform the classification.
ARIIHA enhances in-context learning by retrieving relevant demonstrations adaptively through two distinct strategies: one prioritizes demonstrations that focus on similar target groups, while the other selects those with high similarity scores. Our method dynamically determines the optimal retrieval strategy using thresholds derived from similarity scores and the model’s dependence on shortcut cues, ensuring that only reliable contextual demonstrations contribute to classification.

Extensive experiments on the Implicit Hate Corpus (IHC) dataset \cite{elsherief-etal-2021-latent} demonstrate that our framework consistently outperforms existing state-of-the-art baselines in few-shot learning scenarios for implicit hate speech detection. 
Furthermore, the ablation and case studies demonstrate the overall optimality of our proposed ARIIHA approach.

% To the best of our knowledge, this is the first study to propose a retrieval-based in-context learning approach in an adaptive manner, pioneering a novel methodology for implicit hate speech detection. Our work prioritizes targeted groups in retrieval-based in-context learning and makes three key contributions. 
% \textit{First}, we propose a novel adaptive retrieval-based in-context learning method that dynamically captures decision conditions depending on datasets while mitigating biases influenced by human annotators. 
% \textit{Second}, we address the challenge of over-sensitivity, a common limitation in existing LLM-based hate speech detection models, ensuring more balanced and reliable predictions.  
% \textit{Third}, we conduct extensive experiments on the IHC dataset, demonstrating that our ARIIHA approach consistently outperforms state-of-the-art baselines in few-shot in-context learning for implicit hate speech detection.  

\begin{figure}[!ht]
\centering
\includegraphics[width=0.46\textwidth]{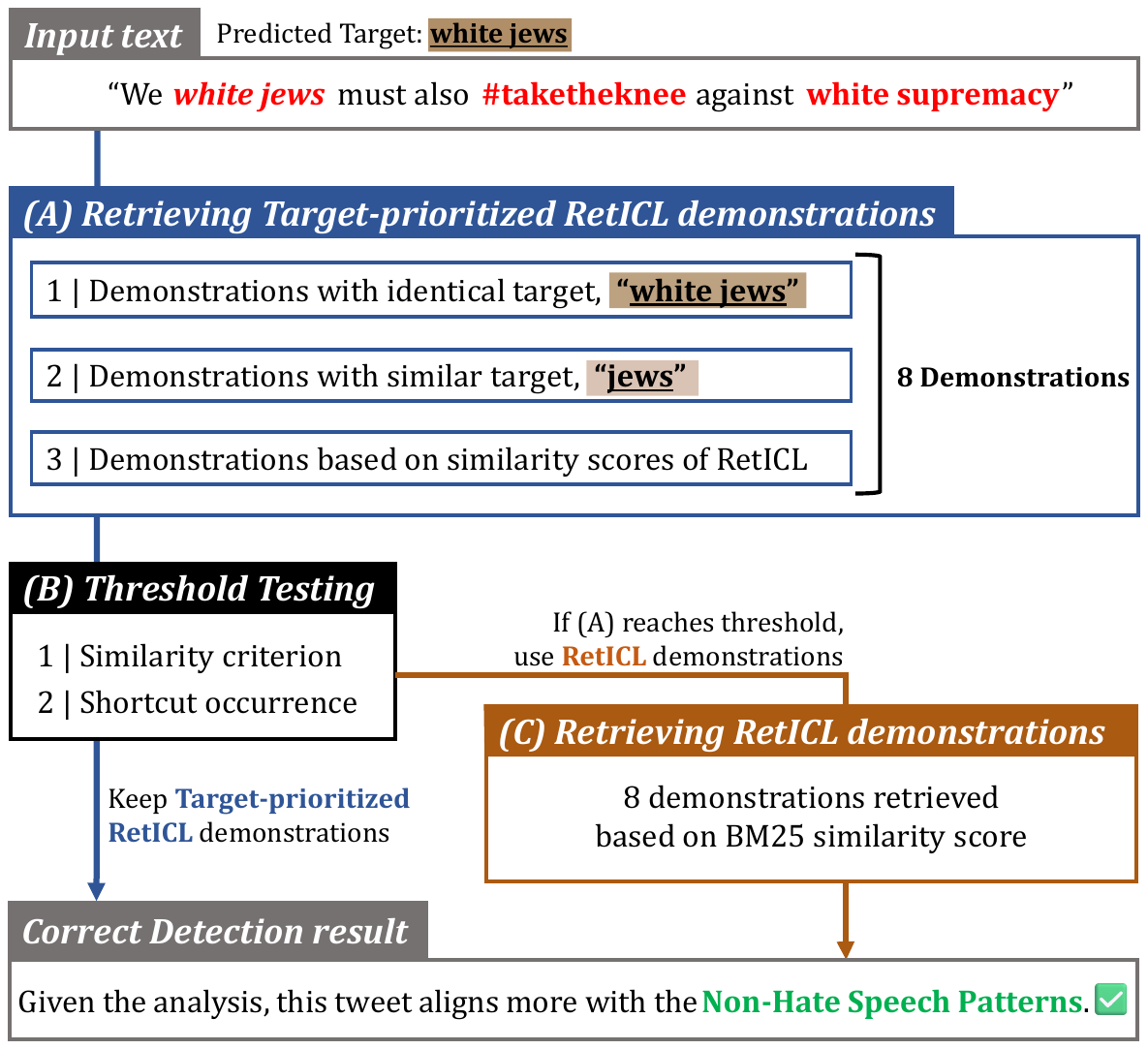}
\vspace{-2.5mm}
\caption{Overall pipeline of our proposed ARIIHA approach. Despite the presence of sensitive words in the input text, colored in \textcolor{red}{red}, ARIIHA accurately detects the correct label while mitigating over-sensitivity.}
%\vspace{-3.5mm}
\vspace{-3.9mm}
\label{fig:method}
\end{figure}

\section{METHOD}

% \begin{figure}[!ht]
% \centering
% % \includegraphics[width=0.95\textwidth]{rsc/fig1_method.pdf}
% \includegraphics[width=0.48\textwidth]{rsc/fig1_method_pipeline_v2.pdf}
% \vspace{-3mm}
% \caption{Overall pipeline of our proposed ARIIHA approach. Despite the presence of sensitive words in the input text, colored in \textcolor{red}{red}, ARIIHA accurately detects the correct label while mitigating over-sensitivity.}
% \vspace{-3.5mm}
% \label{fig:method}
% \end{figure}

Hate speech detection aims to classify text as either hateful or benign automatically. Our approach employs Retrieval-based In-context Learning (RetICL) \cite{luo2024context, wang-etal-2024-learning, li-etal-2023-unified} by incorporating carefully selected demonstrations into the model's prompt, thereby guiding its decision-making process.

\subsection{Retrieval-based In-context Learning}
%We start with retrieval-based in-context learning (RetICL) to enhance implicit hate speech detection. 
Our method begins with general RetICL to enhance implicit hate speech detection.
Given that hate speech detection is highly sensitive to the presence of specific words, we utilize BM25 \cite{robertson1976relevance, robertson2009probabilistic}, a term-based sparse retrieval model, instead of a dense retrieval approach \cite{song2020mpnet}. The sparse retrieval framework is particularly well-suited for this task, as it effectively prioritizes keyword-based relevance, aligning with the linguistic characteristics of hate speech. 
RetICL efficiently retrieves demonstrations based on the top-k highest similarity scores, making it an effective approach. By leveraging a sparse retrieval model, RetICL selects demonstrations primarily based on lexical features, as well as the overall format and tone of the text. 
However, in the hate speech detection task, this approach can sometimes prioritize words that are not crucial for detection or become overly sensitive to specific toxic terms, potentially leading to misclassifications.

\begin{table*}
  \centering
  \caption{Qwen2.5-7B-Instruct results on various In-context Learning demonstration settings.}
  \vspace{-3mm}
  \label{tab:main_result}
  \begin{tabular}{lcccccccc}
    \toprule
    Demonstration & Macro-F1 & Weighted-F1 & Balanced-Acc & Prec@1 & Rec@1 & Prec@0 & Rec@0 & Over-Sensitivity \\
    \midrule
    Random (0-shot) & 69.52 & 70.30 & 71.04 & 58.44 & 76.04 & 81.45 & 66.05 & 17.60 \\
    Random (4-shot) & 70.82 & 71.82 & 71.80 & 60.83 & 73.20 & 80.70 & 70.39 & 12.37 \\
    Random (8-shot) & 70.46 & 71.30 & 71.82 & 59.71 & 75.74 & 81.67 & 67.89 & 16.03 \\
    \midrule
    RetICL (4-shot) & 72.50 & 73.55 & 73.27 & 63.23 & 73.31 & 81.38 & 73.24 & 10.08 \\
    RetICL (8-shot) & 73.76 & 74.77 & 74.54 & 64.68 & 74.71 & 82.40 & 74.37 & 10.03 \\
    \midrule
    Target-prioritized RetICL (4-shot) & 72.55 & 73.66 & 73.20 & 63.69 & 72.30 & 80.98 & 74.10 & \textbf{8.61} \\
    Target-prioritized RetICL (8-shot) & 73.39 & 74.46 & 74.07 & 64.58 & 73.43 & 81.74 & 74.71 & 8.85 \\
    \midrule
    \textbf{ARIIHA (4-shot)} & 75.42 & 76.39 & 76.20 & 66.67 & 76.40 & 83.67 & 75.99 & 9.73 \\
    \textbf{ARIIHA (8-shot)} & \textbf{76.79} & \textbf{77.73} & \textbf{77.54} & \textbf{68.31} & \textbf{77.69} & \textbf{84.69} & \textbf{77.40} & 9.38 \\  
    \midrule
     Upper Bound (4-shot) & 76.79 & 77.74 & 77.48 & 68.54 & 77.23 & 84.46 & 77.73 & 8.69 \\
     Upper Bound (8-shot) & 79.90 & 80.74 & 80.62 & 72.13 & 80.85 & 87.00 & 80.40 & 8.72 \\
    \bottomrule
  \end{tabular}
  \vspace{-2mm}
\end{table*}

\subsection{Target-prioritized RetICL}
Hateful content is directed at specific demographic groups, referred to as targeted groups (e.g., Figure ~\ref{fig:method} \textit{"white Jews"}). Even when texts contain the same or similar targeted groups, their labels may differ depending on the overall meaning and intent. Therefore, retrieving demonstrations with matching or similar targeted groups helps the detection model develop a deeper understanding of the specific decision criteria, rather than being overly influenced by the presence of toxic terms associated with targeted groups. This approach effectively mitigates a key limitation of RetICL, which can sometimes prioritize lexical similarity over contextual meaning, leading to over-sensitivity to toxic words without fully considering the nuanced intent behind each instance.

However, applying this approach to the Implicit Hate datasets presents an additional challenge, as the dataset contains target group annotations but includes NULL values in the target group column. To address this, our proposed Target-prioritized RetICL framework begins by predicting the missing target groups using Qwen2.5-7B-Instruct \cite{yang2024qwen2} with 8-shot in-context learning reasoning. This step ensures that target group information is consistently available, facilitating more robust and generalizable retrieval, particularly for datasets that lack explicit target group annotations.

Once the target group is predicted, the Target-prioritized RetICL framework retrieves demonstrations based on a structured prioritization strategy consisting of three levels:  

\noindent \textbf{1st priority:} Demonstrations that exactly match the predicted target group.  

\noindent \textbf{2nd priority:} Demonstrations that contain a similar target group.  

\noindent \textbf{3rd priority:} Demonstrations retrieved using BM25, ranked by high similarity scores.  

To ensure optimal retrieval, we prioritize high-priority examples within the top-k limit. If an insufficient number of first-priority demonstrations is available, we include all available first-priority examples and then maximize the inclusion of second-priority demonstrations within the remaining capacity. If neither first- nor second-priority demonstrations are present in the retrieval pool, we allocate the remaining space to third-priority examples. 
For instance, in Figure~\ref{fig:method} (A), the retrieval process begins by selecting one demonstration from the first priority category, which contains an identically matching targeted group, \textit{"white Jews"}. Next, for the second priority, BM25 retrieves two demonstrations with similar targeted groups, specifically those targeting \textit{"Jews"}. Finally, to fill the remaining capacity, five additional demonstrations with the highest similarity scores are selected.

Despite its robustness, Target-prioritized RetICL still has certain limitations. By prioritizing demonstrations that target the same or similar groups, the retrieved demonstrations may have lower similarity scores compared to those selected by RetICL. Additionally, this approach may overlook important lexical patterns that serve as key indicators for accurate detection, potentially affecting overall performance. Furthermore, if the model fails to accurately predict the targeted group of the input text, the entire pipeline risks collapsing at the initial step, highlighting the necessity of an adaptive threshold to ensure a stable and consistent retrieval process.
\vspace{-2.5mm}

\begin{table}
  \centering
  \caption{Qwen2.5-7B-Instruct results of RetICL 8-shot.}
  \vspace{-3mm}
  \label{tab:retrieval}
  \begin{tabular}{lcccc}
    \toprule
    Retriever & Macro-F1 & Bal-Acc & Prec@1 & Rec@1 \\
    \midrule
    % all-mpnet-base-v2  & 66.09 & 69.39 & 52.85 & 81.74 \\
    all-mpnet-base-v2  & 72.25 & 72.79 & 63.64 & 71.09 \\
    BM25  &  \textbf{73.76} &  \textbf{74.54} & \textbf{64.68} & \textbf{74.71}  \\
    \bottomrule
    \vspace{-10mm}
  \end{tabular}
\end{table}

\subsection{ARIIHA: Adaptive RetICL}
To dynamically leverage the strengths of both RetICL and Target-prioritized RetICL while addressing their respective limitations, Adaptive Retrieval-based In-context Learning (ARIIHA) optimally selects demonstrations from both approaches. This method enhances the Macro-F1 score while mitigating over-sensitivity in implicit hate speech detection.  

The overall framework is illustrated in Figure~\ref{fig:method}. To enable adaptive selection, we introduce two thresholds: one based on the number of retrieved demonstrations that fail to meet the BM25 similarity criterion, and the other on whether a shortcut occurs, directly influencing the detection result. If the retrieved demonstrations from Target-prioritized RetICL reach both thresholds, they are replaced with demonstrations from RetICL.

Specifically, the replacement follows these conditions:  
(1) If the BM25 similarity score of the demonstrations retrieved from Target-prioritized RetICL is equal to or lower than the predefined threshold, and  
(2) If the classification model reaches the final detection decision through a shortcut, relying on only 1-to-3 word phrases enclosed in quotation marks (" ") without deeper contextual reasoning.

For condition (1), we determine the optimal similarity score threshold using a nested loop algorithm, iterating through the similarity score range from 0 to 150. The threshold that yields the highest Macro-F1 score is selected as the optimal similarity score threshold. Target-prioritized demonstrations are replaced with RetICL demonstrations only if both conditions (1) and (2) are met. Conversely, if the retrieved Target-prioritized demonstrations do not satisfy at least one of the conditions, they are retained in the final selection.

For example, in Figure ~\ref{fig:method} (B), we first perform threshold testing on demonstrations retrieved from (A) Target-prioritized RetICL. Since the similarity scores of the two demonstrations fall below the optimal threshold, the input text instead retrieves demonstrations from (C) RetICL. The highest similarity demonstrations are selected, including examples such as \textit{"white supremacy is a crime against humanity"}. Ultimately, despite the presence of sensitive words in the input text, such as \textit{"white Jews"}, \textit{"\#taketheknee"}, and \textit{"white supremacy"}, the use of adaptively selected demonstrations enables the model to achieve the correct detection result.

\section{EXPERIMENTS}

\subsection{Experimental Setup}
We utilize the Implicit Hate datasets for our experiments. The retrieval pool consists of the training set of the IHC dataset, while the threshold is optimized on the dev set. The final evaluation is then conducted on the test set.
Since our overall pipeline employs BM25, a sparse retrieval model, rather than a dense retrieval approach, we first compare their effectiveness in hate speech detection tasks. The performance of each retrieval model—sparse and dense—within the RetICL framework is presented in Table~\ref{tab:retrieval}.

\subsection{Experimental Results}
The performance of ARIIHA compared to various in-context learning settings is presented in Table~\ref{tab:main_result}. Over-sensitivity is quantified by subtracting the Precision score of the 'hate' label from its Recall score, following the approach of \cite{zhang-etal-2024-dont-go}.  The results indicate that few-shot learning with randomly selected demonstrations yields minimal improvement. However, ARIIHA demonstrates a significant performance gain, surpassing both randomly selected demonstrations and the individual RetICL and Target-prioritized RetICL approaches.  

Table~\ref{tab:baseline_comparison} presents a comparative analysis of ARIIHA against various hate speech detection baselines that employ supervised fine-tuning and in-context learning. The results demonstrate that retrieval-based in-context learning not only consistently outperforms standard in-context learning approaches but also surpasses certain supervised fine-tuning methods. 
Notably, ARIIHA achieves superior performance, exceeding CoT reasoning \cite{yang-etal-2023-hare} by a substantial margin of over 30\%, highlighting the effectiveness of adaptive demonstration selection in improving hate speech detection. Furthermore, the small performance gap of 3.11\% between ARIIHA and the upper bound—where either Target-prioritized RetICL or RetICL succeeds, except when both lead to misclassification—indicates that ARIIHA’s adaptive threshold is successfully optimized, effectively balancing precision and recall.

\begin{table}
  \caption{Qwen2.5-7B-Instruct results on various hate speech detection baselines.}
  \vspace{-2mm}
  \label{tab:baseline_comparison}
  \begin{tabular}{llcc}
    \toprule
    Method & Model & F1 & Accuracy \\
    \midrule
    \multicolumn{4}{c}{\textbf{Supervised Fine-Tuning (SFT)}} \\
    \midrule
    Co-HARE & GPT-2 large & 71.58  & 80.07 \\
    % Co-HARE & Flan-T5 small & 73.49 & 78.54 \\
    % Co-HARE & Flan-T5 base & 75.69 & 80.38 \\
    Co-HARE & T5 large & 75.88 & 80.98 \\
    Co-HARE & Flan-T5 large & \textbf{76.71} & \textbf{81.49} \\
    
    % C & GPT-2 large & 66.68 & 73.32 \\
    % C+T+I & GPT-2 large & 65.25 & 75.95 \\
    % Fr-HARE & GPT-2 large & 71.35 & 78.47 \\
    % Co-HARE & GPT-2 large & \textbf{71.58} & \textbf{80.07} \\
    \midrule
    \multicolumn{4}{c}{\textbf{In-Context Learning}} \\
    \midrule
    CoT & QWEN2.5-7B & 42.39 & 62.15 \\
    Vanilla QA &QWEN2.5-7B  & 43.86 & 69.02 \\
    Target & QWEN2.5-7B & 50.40 & 69.53 \\
    Choice QA & QWEN2.5-7B & 52.01 & 69.41 \\
    Cloze & QWEN2.5-7B & \textbf{65.66} & \textbf{64.91} \\
    \midrule
    \multicolumn{4}{c}{\textbf{Retrieval-based In-Context Learning (8-shot)}} \\
    \midrule
    RetICL &  QWEN2.5-7B  & 69.34 & 74.51 \\
    Target-prioritized RetICL &  QWEN2.5-7B  & 68.72 & 74.21 \\
    % \midrule
    % \multicolumn{4}{c}{\textbf{Adaptive Retrieval-based In-Context Learning (Ours)}} \\
    % \midrule
    % ARIIHA (4-shot) &  QWEN-2.5-7B  & 71.20 & 76.15 \\
    \textbf{ARIIHA (Ours)} &  QWEN2.5-7B  & \textbf{72.70} & \textbf{77.51} \\
    \bottomrule
  \end{tabular}
  \vspace{-2mm}
\end{table}

\subsection{Ablation Studies}
\begin{figure}[t]
\centerline{\includegraphics[scale=0.22]{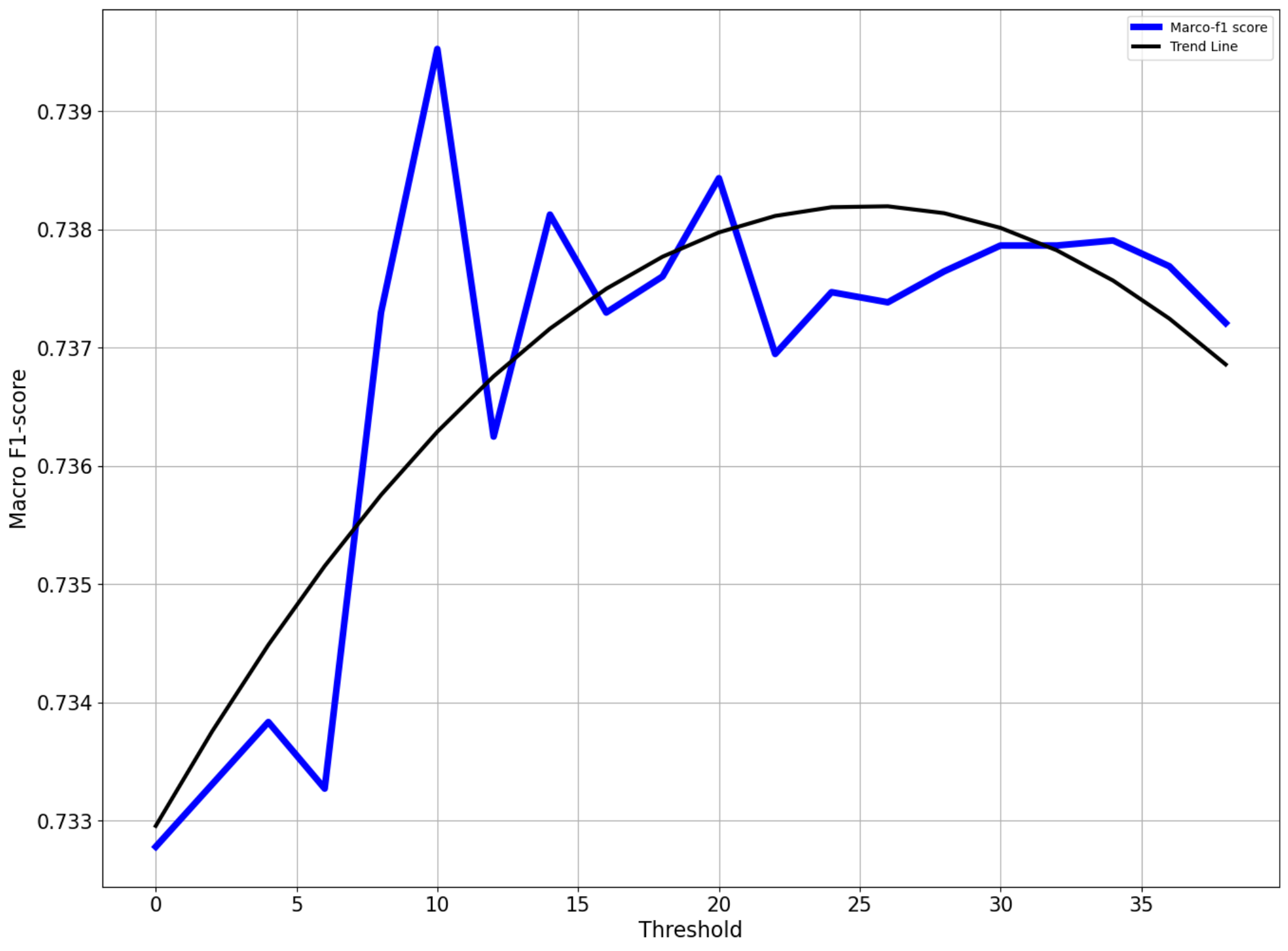}}
\caption{Performance graph of Qwen2.5-7B-Instruct across varying BM25 similarity score thresholds, with an optimal threshold identified at 10.}
\vspace{-5mm}
\label{fig:optimal_threshold}
\end{figure}

We conduct an ablation study to evaluate the necessity of each component in our method and the determination of the optimal threshold, as presented in Table~\ref{tab:demo_combination} and Figure~\ref{fig:optimal_threshold}, respectively.  

Table~\ref{tab:demo_combination} demonstrates that simply mixing retrieved demonstrations from two fixed settings is not suitable for the sensitive nature of the hate speech detection task. Instead, determining the optimal threshold requires a comprehensive evaluation of multiple factors, including the target group of the input text, the presence of similar target examples in the retrieval pool, their similarity scores, and the reasoning process of the detection model—particularly whether a shortcut occurs. The results indicate that removing either the similarity threshold or the shortcut threshold leads to a significant performance drop, comparable to excluding both. This underscores the effectiveness and robustness of our proposed thresholding approach, which evaluates both criteria to optimize retrieval-based in-context learning for hate speech detection. Furthermore, Figure~\ref{fig:optimal_threshold} illustrates the relationship between the similarity score threshold and the Macro-F1, forming a convex curve. This finding highlights the importance of identifying an optimal threshold, as it plays a critical role in enhancing detection performance.

\begin{table}
  \centering
\caption{Ablation study on the impact of two thresholds using Qwen2.5-7B-Instruct in an 8-shot learning setting. M-F1 is the Macro-F1 score, while P and R represent the Precision and Recall for the 'hate' label.}
\vspace{-2mm}
  \label{tab:demo_combination}
  \begin{tabular}{lcccc}
    \toprule
    Threshold & M-F1 & Bal-Acc & P & R \\
    \midrule
    ARIIHA (Ours) & 76.79 & 77.54 & 68.31 & 77.69 \\
    w/o similarity & 73.75 & 74.46 & 64.88 & 74.06 \\
    w/o shortcut & 73.88 & 74.60 & 64.99 &74.30 \\
    w/o similarity and shortcut & 73.39 & 74.07 & 64.58 & 73.43 \\
    \bottomrule
  \end{tabular}
\vspace{-3mm}

\end{table}

\subsection{Case Study}

% \begin{table}[ht]
% \centering
% \caption{Case study on success and failure of ARIIHA.}
% % \vspace{-1mm}
% \small
% \scalebox{0.95}{
% \begin{tabularx}{\linewidth}{lX} % Added "l" for proper alignment
% \hline
% Case A \\
% \textbf{Input Text} & \colorbox{cyan}{hindus} sikhs in afghanistan feel \textcolor{red}{neglected discriminated} - \#hdl [\textcolor{green}{not hate}]\\  
% \hline
% \textbf{Demonstration} & the grand \colorbox{cyan}{hindus} alliance of narendra modi is likely to deliver 350 seats in parliament 2014. nitish repents. [\textcolor{green}{not hate}] \\  
% \hline
% \textbf{Correct} & The tweet expresses concern and support for a specific community facing discrimination. The language used is neutral and focuses on the experiences of a particular religious minority.  
% Therefore, this tweet is \textcolor{green}{not hate speech}. \\  
% \hline
% \hline
% \textbf{Input Text} &i have even less sympathy for \colorbox{cyan}{refugees} now [\textcolor{red}{hate}]\\  
% \hline
% \textbf{Demonstration} & weird how they tolerate sharia  but not white nationalism  or even \colorbox{cyan}{white refugees}. [\textcolor{red}{hate}]\\  
% \hline
% \textbf{Incorrect} & 

% The tweet is a general statement expressing a personal opinion about refugees. It does not promote any form of discrimination or xenophobia, so \textcolor{green}{not hate speech}.
% \\  
% \hline
% \end{tabularx}
% }
% \label{tab:case_study_example}
% \vspace{-4mm}
% \end{table}

\begin{table}[ht]
\centering
\caption{Case study on the success and failure of ARIIHA.}
\small
\scalebox{0.97}{
\begin{tabularx}{\linewidth}{lX} 
\hline
\multicolumn{2}{c}{\textbf{Case A}} \\  % Centered across two columns
\hline
\textbf{Input Text} & \colorbox{cyan}{hindus} sikhs in afghanistan feel \textcolor{red}{neglected discriminated} - \#hdl [\textcolor{green}{not hate}]\\  
\hline
\textbf{Demonstration} & the grand \colorbox{cyan}{hindus} alliance of narendra modi is likely to deliver 350 seats in parliament 2014. nitish repents. [\textcolor{green}{not hate}] \\  
\hline
\textbf{Correct} & The tweet expresses concern and support for a specific community facing discrimination. The language used is neutral and focuses on the experiences of a particular religious minority.  
Therefore, this tweet is \textcolor{green}{not hate speech}. \\  
\hline
\hline
\multicolumn{2}{c}{\textbf{Case B}} \\  % Another case (if needed)
\hline
\textbf{Input Text} & I have even less sympathy for \colorbox{cyan}{refugees} now [\textcolor{red}{hate}]\\  
\hline
\textbf{Demonstration} & Weird how they tolerate sharia but not white nationalism or even \colorbox{cyan}{white refugees}. [\textcolor{red}{hate}]\\  
\hline
\textbf{Incorrect} & The tweet is a general statement expressing a personal opinion about refugees. It does not promote any form of discrimination or xenophobia, so it is classified as \textcolor{green}{not hate speech}. \\  
\hline
\end{tabularx}
}
\label{tab:case_study_example}
\vspace{-4mm}
\end{table}

Table~\ref{tab:case_study_example} presents both a success and a failure case of ARIIHA. In both cases, Target-prioritized demonstrations were replaced with RetICL demonstrations due to their low similarity scores. The input texts contain sensitive terms such as \textit{"neglected discriminated"} and \textit{"refugees"}. Case B illustrates a failure case where the model fails to correctly capture the implicit hateful intent, as it may not always be sufficient to recognize implicit hate intention and can lead to misinterpretations as personal opinions. However, as demonstrated in Case A, ARIIHA effectively adapts to the dataset by dynamically selecting relevant demonstrations, ultimately improving detection performance in general cases.

\section{CONCLUSION}
Implicit hate speech detection remains a challenging task in natural language processing. To address the over-sensitivity and subjectivity issues present in previous detection methods, we introduce \textbf{ARIIHA}, a novel approach that adaptively retrieves a small set of demonstrations targeting the same or similar groups. 
This study is the first to propose a \textbf{Retrieval-based In-context Learning} framework for implicit hate speech detection. Experimental results demonstrate that ARIIHA outperforms state-of-the-art in-context learning baselines, highlighting its effectiveness in improving detection accuracy while mitigating over-sensitivity and dataset-dependent subjectivity.

% \begin{acks}
% \end{acks}

\bibliographystyle{ACM-Reference-Format}
\bibliography{A_citation}
\end{document}